%% file: emnlp_final.tex
\pdfoutput=1
\documentclass[11pt]{article}
\usepackage[final]{acl}
\usepackage{comment}
\usepackage{tabularx} % Add this in your preamble
\usepackage{times}
\usepackage{latexsym}
\usepackage{array}
\usepackage{subcaption}
\usepackage{pgfplots}
\usepackage{geometry}
\usepackage{longtable}
\usepackage{float}
\usepackage[T1]{fontenc}
\DeclareUnicodeCharacter{1EBF}{\'{e}}   % ế
\DeclareUnicodeCharacter{1EA1}{\d{a}}   % ạ
\DeclareUnicodeCharacter{1EE7}{\u{u}}   % ủ
\DeclareUnicodeCharacter{1ED9}{\d{o}}   % ộ
\DeclareUnicodeCharacter{1EAB}{\~{a}}   % ẫ
\DeclareUnicodeCharacter{1EC3}{\~{e}}   % ễ
\DeclareUnicodeCharacter{1ED1}{\^{o}}   % ố
\DeclareUnicodeCharacter{1EEF}{\~{u}}   % ữ
\DeclareUnicodeCharacter{0126}{\H{H}} % uppercase Ħ
\DeclareUnicodeCharacter{0127}{\H{h}} % lowercase ħ
\usepackage[utf8]{inputenc}

\usepackage{amssymb}  % For general symbols (optional but common)
\usepackage{pifont}   % For checkmark and xmark
\newcommand{\Checkmark}{\ding{51}}     
\newcommand{\XSolidBrush}{\ding{55}}  
\usepackage{microtype}

\usepackage{inconsolata}
\usepackage{graphicx,subcaption,booktabs,float}

\usepackage[english, arabic,hindi,greek]{babel}
\usepackage{devanagari} 
\usepackage{microtype}
\usepackage{amsmath}

\usepackage{multirow}
\usepackage{times}
\usepackage{latexsym}
\usepackage{subcaption}
\usepackage{rotating}

\usepackage{xcolor}
\usepackage{amssymb}
\title{Tokenization and Representation Biases in Multilingual Models on Dialectal NLP Tasks}
\author{
\textbf{Vani Kanjirangat\textsuperscript{1}},
\textbf{Tanja Samardžić\textsuperscript{1}},
\textbf{Ljiljana Dolamic\textsuperscript{2}},
\textbf{Fabio Rinaldi\textsuperscript{1}} \\[0.5em]
\textsuperscript{1}SUPSI, IDSIA, Switzerland\\
\textsuperscript{2}armasuisse S+T, Switzerland\\[0.5em]
\small
\href{mailto:vani.kanjirangat@supsi.ch,tanja.samardzic@supsi.ch,fabio.rinaldi@supsi.ch}{\{vani.kanjirangat, tanja.samardzic, fabio.rinaldi\}@supsi.ch}, 
\href{mailto:Ljiljana.Dolamic@armasuisse.ch}{Ljiljana.Dolamic@armasuisse.ch}
}

\begin{document}

\selectlanguage{english}
%\setmainfont{Noto Sans Devanagari}
\maketitle
\begin{abstract}
Dialectal data are characterized by linguistic variation that appears small to humans but has a significant impact on the performance of models. This dialect gap has been related to various factors (e.g., data size, economic and social factors) whose impact, however, turns out to be inconsistent. In this work, we investigate factors impacting the model performance more directly: we correlate Tokenization Parity (TP) and Information Parity (IP), as measures of representational biases in pre-trained multilingual models, with the downstream performance.  
We compare state-of-the-art decoder-only LLMs with encoder-based models across three tasks: dialect classification, topic classification, and extractive question answering, controlling for varying scripts (Latin vs. non-Latin) and resource availability (high vs. low). Our analysis reveals that TP is a better predictor of the performance on tasks reliant on syntactic and morphological cues (e.g., extractive QA), while IP better predicts performance in semantic tasks (e.g., topic classification).
Complementary analyses, including tokenizer behavior, vocabulary coverage, and qualitative insights, reveal that the language support claims of LLMs often might mask deeper mismatches at the script or token level\footnote{Code at \url{https://github.com/vanikanjirangat/Tokenizer\_Fairness\_Dialect}}.
\end{abstract}

\section{Introduction}\label{sec:intro}

\selectlanguage{english}
Large Language Models (LLMs) pre-trained on massive text data in many languages have become the preferred solution for various Natural Language Processing (NLP) tasks. The use of this technology for processing dialects and regional varieties remains limited. Small variations in pronunciation and writing \cite{zampieri2018language, scherrer2023tenth, habash2024proceedings}, which humans can easily ignore, lead to significant performance drops known as the \textit{dialect gap} \cite{kantharuban-etal-2023-quantifying}. Including this variation is hard, although important for more human-like interactions with LLMs \cite{amadeus-etal-2024-bridging}. It is especially important for a wide linguistic coverage, as many languages are not standardized or have multiple standards \cite{Samardzic-Ljubesis-2021}. 

In previous studies, dialect variances have been related to economic and social factors \cite{kantharuban-etal-2023-quantifying}, but the effects were inconsistent across different settings. Looking for more consistent factors directly related to how LLMs work, we turn to the representational biases in multilingual LLMs. 

We study two aspects where the biases can be quantified with recently proposed measures. First, the tokenization bias has been shown to impact not only the performance, but also the costs of deploying LLMs across languages \cite{ahia2023all}. Recently, this bias was quantified as Tokenization Parity (TP) \cite{petrov2024language}.  Second, Information Parity (IP) \cite{tsvetkov2024information}  measures how well an LLM compresses or represents the same content across languages. In both cases, the measures show a difference between a given language and English as a reference language. 

To address the dialect gap, we correlate these measures to downstream performance on three dialect NLP tasks, each targeting a different level of representation: Dialect Identification (DI), which mostly relies on surface-level clues, Topic Classification (TC) as a primarily semantic task, and Extractive Question Answering (EQA) as a task that relies on both kinds of features. In all three cases, we work with multiple data sets representing different economic and cultural settings. This allows us to control for additional factors that are known to play an important role in creating biases. In particular, we control for the script (Latin vs. non-Latin) and resource level (high vs. low) \cite{van2022writing}. On the model side, we control for the general type of pre-trained LLMs, distinguishing between encoder-only (BERT-based) and decoder-only (E.g., GPT) multilingual models. 

The \textbf{Key Findings} are:
\begin{enumerate}
\setlength\itemsep{-0.5em}
\item Encoder-based models consistently outperform decoder-only LLMs\footnote{All models are referred to as LLMs and are distinguished as encoder-only or decoder-only where relevant} across the evaluated dialectal tasks. 
\item TP is more sensitive to the type of the script, while IP reflects biases influenced by both script and resource availability. Additionally, both metrics show model-dependent variation, highlighting how architectural and training differences contribute to representational disparities.
\item Information Parity (IP) shows more substantial alignment with tasks requiring semantic understanding and complex reasoning, while Tokenization Parity (TP) is more predictive for tasks that rely on morphological and syntactic features, especially span-based extractive tasks. These correlations are further modulated by language resource availability and script type.
\end{enumerate}
\begin{figure*}
    \centering
    \includegraphics[width=1\linewidth]{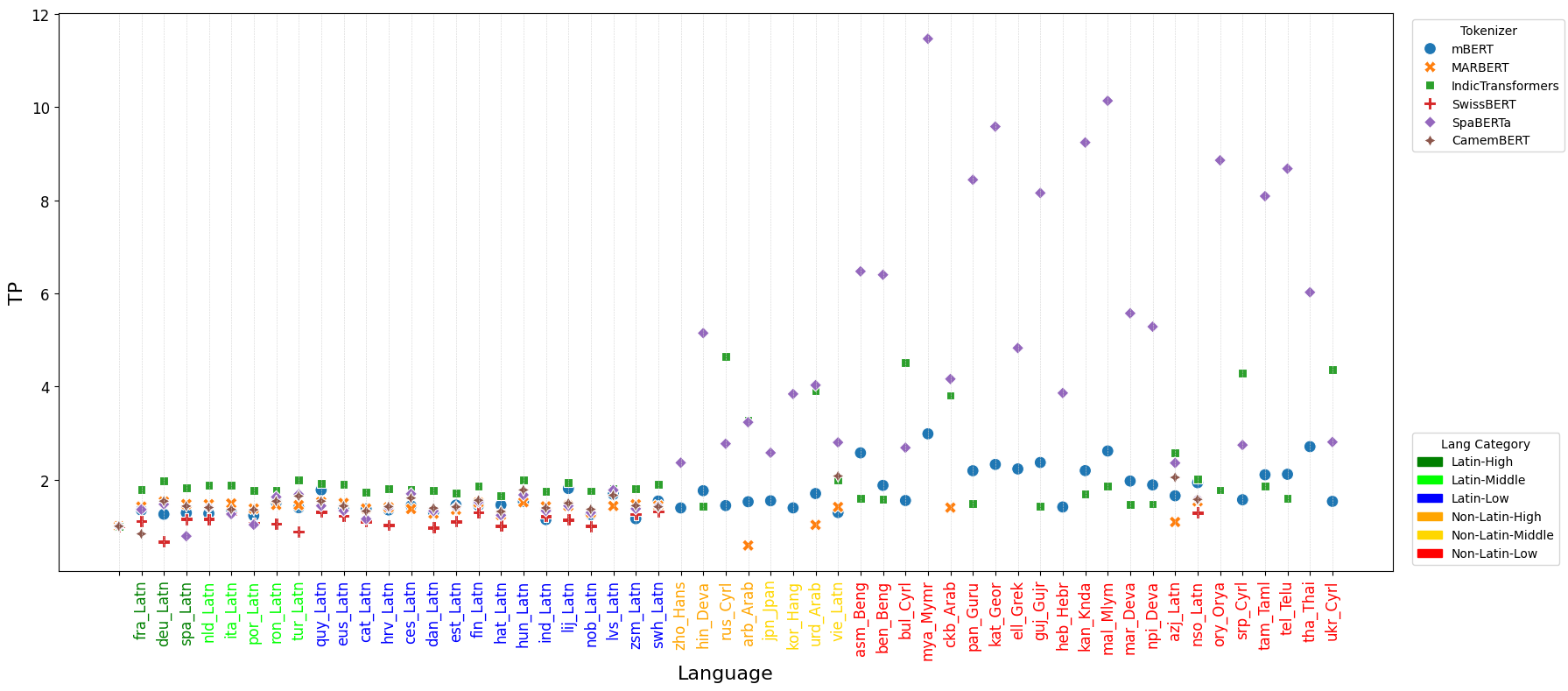}
    \caption{Tokenization Parity (TP) across languages resulting from the tokenizers used in encoder-type models.}
    \label{fig:TP_Enc_nw}
\end{figure*}
\section{Dialect Tasks, Data and Models}
\label{sec:data}

Our selection of dialect NLP tasks, data and models was guided by the goal of covering as diverse settings as possible while keeping the computation feasible. 

\subsection{Tasks}

\paragraph{Dialect Identification (DI)} This task consists of assigning a dialect or region label to each input sentence or utterance. This task comes in two versions: monolabel (each utterance can belong to only one dialect) and multilabel (some utterances can belong to multiple dialects), with the latter being more realistic but harder to perform and evaluate. We used the datasets from several VarDial shared tasks: Nuanced Arabic Dialect Identification (NADI-2023), Swiss German dialect identification (GDI), Indo-Aryan Language Identification (ILI), and multi-label DSL-ML datasets \cite{abdul2023nadi,samardzic2016archimob,zampieri2018language,chifu2024vardial}. 

\paragraph{Topic Classification (TC)} This task is similar to the monolabel DI task in that each snippet of text is assigned a single topic label. The difference is that predicting the label requires neutralizing surface-level differences between dialects. This task is included in the DialectBench benchmark \cite{faisal2024dialectbench} as the SIB-200 dataset   \cite{adelani2024sib} representing 200 languages. The topic classes include: \{science/technology travel, politics, sports, health, entertainment, geography\}. We conducted the fine-tuning experiments on 29 languages belonging to different scripts, along with the availability of language resources: eight Latin-high, nine Latin-low, five non-Latin-high, and seven non-Latin-low. 

\paragraph{Extractive Question Answering (EQA)} This task combines in some way the features of the previous two, as it requires identifying the relevant spans (surface features) but relying on deeper semantic representation (understanding the question-answer relationships). We experimented on 24 dialectal variants - eleven Latin-high, two Latin-low, nine non-Latin-High, and two non-Latin-low, from the dataset  SDQA \cite{faisal2021sd} also provided via the DialectBench.  

%\bigskip
%\noindent
The general statistics and further details of datasets are given in Appendix \ref{app:data}. The class distributions for DI and TC tasks are shown in Figures \ref{fig:Class_distribution_DI}, \ref{fig:Class_distribution_TC} in Appendix \ref{app:data}.
% The details of the script and resource categorizations are presented in Table \ref{tab:language_groups} in Appendix \ref{app:category_details}.

\section{The Bias Metrics}\label{sec:bias-metrics}
% : Tokenization Parity (TP) \&  Information Parity (IP)
\subsection{Models \& Tokenizers} 
\label{sec:models}
\paragraph{Encoder type} 
We used the multilingual mBERT for the encoder variant in all the tasks. Specifically, in the case of the DI task, we performed additional comparisons between mBERT and language-specific variants such as MARBERT (Arabic), IndicBERT(Indic), German-BERT\footnote{We also did experiments with Swiss-BERT, which gave similar performance as German-BERT} (Swiss-German), SpanBERTa (Spanish), and CamemBERT (French). We use the respective models from HuggingFace\footnote{\url{https://huggingface.co/}}. 

\paragraph{Decoder type}
Among the multilingual decoder-type models, we selected Phi-3.5-mini, Llama 3.2-3B, Mistral-7B, Falcon-7B, Gemma-7B, and SILMA-9B models. SILMA-9B represents an Arabic-specific LLM, while the other models discussed are English-centric or generalized multilingual LLMs, claiming support for a broad spectrum of languages. For the downstream task performance evaluation, we considered Phi-3.5 and Llama-3.2 by supervised fine-tuning (SFT) experiments to compare with the encoder variants.  

% \vspace{-5mm}
\paragraph{Tokenizer \& Languages:} 
Language and tokenization are deeply intertwined, shaping LLMs’ multilingual capabilities. Most current models use subword tokenization strategies such as BPE, SentencePiece, or byte-level methods. Newer models like LLaMA and Phi adopt the OpenAI tiktoken tokenizer\footnote{\url{https://github.com/openai/tiktoken}}, which operates at the byte level using UTF-8 encoding. This approach is language-agnostic, breaking input into bytes or fragments when unknown tokens are encountered. In contrast, SentencePiece typically defaults to character-level segmentation. Non-Latin scripts (e.g., Arabic, Hindi, Bengali) involve multi-byte characters in UTF-8, making them more prone to token fragmentation under byte-level fallback. This behavior impacts the vocabulary coverage and can hinder effective representation of non-Latin text. Details on tokenizer configurations and vocabulary sizes are provided in Table \ref{tab:tok}, Appendix \ref{app:tok_details}. The fairness or, inversely, the biases of pre-trained multilingual models can be measured considering either the surface level or deeper semantic features. 

\begin{figure*}
    \centering
    \includegraphics[width=1\linewidth]{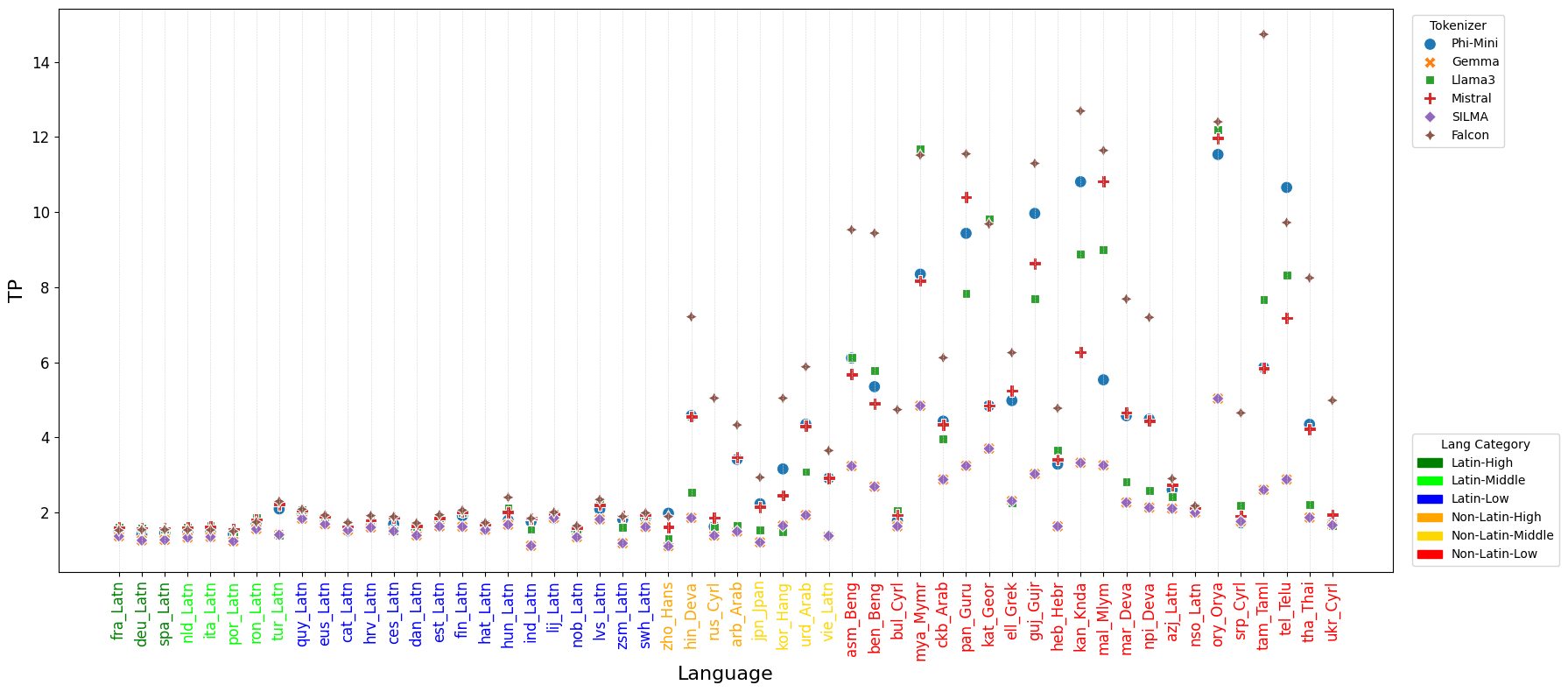}
    \caption{Tokenization Parity (TP) across languages resulting from the tokenizers used in decoder-type models.}
    \label{fig:TP_LLM}
\end{figure*}
\paragraph{Tokenization Parity} Following \citet{petrov2024language}, we use TP as a metric to analyze the tokenization fairness. The metric systematically assesses how well the tokenizers treat parallel sentences across different languages. Parity occurs when a tokenizer exhibits similar tokenized lengths for the same sentence in different languages. Consider a sentence $s_A$ in language A and its translation $s_B$ to language B. Then, a tokenizer \emph{t} achieves parity for A with respect to B at $s_A$ and $s_B$ if  $|t(sA)|/|t(sB)| \approx 1 $, where \(t(s_A)\) is the tokenization of the sentence sA and \(|t(s_A)|\) represents its length. The premium for A relative to B is the ratio $|t(s_A)|/|t(s_B)|$ \cite{petrov2024language}. A value close to 1 indicates fewer splits into subwords, which indicates that the tokenizer vocabulary covers the language well. When the value is greater than 1, it indicates that the language tokenizer requires more tokens to represent the same content. This may indicate a suboptimal representation of the language by the LLM, leading to inefficient representation and potentially poorer downstream task performance. At the same time, these values are language-dependent, and hence, the number of tokens required to represent the same sentence in different languages can affect TP values.

\paragraph{Information Parity} Following \citet{tsvetkov2024information}, we adopt Information Parity (IP) as another metric for evaluating multilingual, specifically dialectal fairness in large language models (LLMs). IP draws on information-theoretic principles and quantifies the LLM's efficiency in compressing text in a given language relative to a reference language. For a text in language \emph{L}, IP is defined as the ratio between the negative log-likelihood of the text in English and the negative log-likelihood of the same text in language L. In this context, English serves as a language-agnostic reference compressor. IP expresses the total amount of information or uncertainty in a sequence perceived by the LLM relative to the reference language. Unlike similar metrics such as perplexity, IP is less sensitive to variations in tokenization across languages and models.

\section{Experiments}\label{sec:eval_methods}
% \begin{table*}
% \centering
% %\renewcommand{\arraystretch}{0.7}
% \scalebox{0.85}{
% \begin{tabular}{p{3.9cm}p{1.5cm}p{1.6cm}p{2cm}p{2cm}p{3cm}}
% \toprule
% \textbf{Language (Type)} & \multicolumn{2}{c}{\textbf{Decoder-only}} & \multicolumn{3}{c}{\textbf{Encoder-only}} \\
% \cmidrule(l){2-3}
% \cmidrule(l){4-6}
% & Phi-3.5 & Llama 3.2 & BERT-base & mBERT & Language-Specific \\
% \midrule
% Arabic(ML) & 0.54 & 0.26 & 0.47  & 0.62 & 0.84(MARBERT) \\
% Swiss-German(ML) & 0.49 & 0.46 & 0.62 & 0.59 & 0.60(Swiss-BERT) \\
% Indo-Aryan(ML) & 0.74 & 0.32 & 0.81 & 0.88 & 0.90(IndicTransformers) \\
% French(MuL) & 0.61 & 0.35 & 0.67 & 0.70 & 0.75(CamemBERT) \\
% Spanish(MuL) & 0.40 & 0.79 & 0.81 & 0.83 & 0.85(spanBERTa) \\
% French(MuLO) & 0.61 & 0.35 & 0.36 & 0.28 & 0.44(CamemBERT) \\
% Spanish(MuLO) & 0.53 & 1.00 & 0.85 & 0.78 & 0.89(spanBERTa) \\
% \bottomrule
% \end{tabular}}
% \caption{\label{tab:Results} Performance on dialect identification task across models. Language tags indicate task type: \textbf{ML} = Mono-label, \textbf{MuL} = Multi-label, \textbf{MuLO} = Multi-label Only.}
% \end{table*}

Our first goal is to evaluate the LLMs' performance on the dialectal downstream tasks, controlling for various factors, namely scripts (Latin vs. non-Latin) and resource levels (High vs. Low). The latter categorization can be slightly biased as sometimes the distinction between high, medium, and low resources can be fine-lined. The categorization is reported in Table \ref{tab:language_groups} of Appendix \ref{app:category_details}. We then quantitatively analyze the model's script and representation biases, measuring the correlation between the observed performance on one side and the two bias metrics --- tokenization parity (TP) and information parity (IP) --- on the other. We complement these analyses with a vocabulary analysis and a manual inspection of the model tokenizers' output.  
% \begin{figure*}
%     \centering
%     \includegraphics[width=1\linewidth]{TP_dec_enlarged.png}
%     \caption{Tokenization Parity (TP) across languages resulting from the tokenizers used in decoder-type models.}
%     \label{fig:TP_LLM}
% \end{figure*}

\subsection{Model Fine-Tuning Methods and Parameters} 
We performed supervised fine-tuning (SFT) of the decoder-only LLMs - Phi-3.5 and Llama 3.2 models and compared them with encoder-only models, mainly mBERT, on the datasets described in Section \ref{sec:data}. We decided to select fewer representative models to economize computing time. On the other hand, the parity score does not require a lot of computation, so we decided to keep multiple models to have a better overview. For decoder-only LLMs' fine-tuning, we used the parameterization techniques (PEFT) \cite{ding2023parameter} with LoRA (Low-Rank Adaptation) \cite{hu2021lora} and bit quantizations to cope with memory issues and efficiency. Four-bit quantizations with LoRA \emph{R=16 or 8} and \emph{alpha =8}, \emph{drop-out = 0.1}, \emph {batch\_sizes = 1, 2 or 4} with \emph{gradient accumulation = 8}, learning rate \emph{lr= 2e-4 or 5e-5} and lr scheduler, mostly cosine else linear were used. Parameter optimizations were done using the hyperparameter optimization framework, Optuna\footnote{\url{https://optuna.org/}}. Further details of general experimental settings can be found in Appendix \ref{app:exp}. The prompts for instruction tuning each task are reported in Appendix \ref{app:prompts}. 

For the encoder-only models, we used full-finetuning (FFT), with 3 epochs of training, AdamW optimizer with learning rate of 2e-5, batch size of 8 or 16, and weight decay of 0.01.

In the multi-label setup of the DI task, we created a representative train-test sample dataset for the French dataset. This reduced the size of this automatically curated dataset (details in Appendix \ref{app:data}) allowing us to avoid unnecessary computing costs. We used a custom trainer function to compute the multi-label loss using Binary Cross-Entropy with Logits (BCELoss with Logits).
\begin{figure*}
    \centering
    \includegraphics[width=1\linewidth]{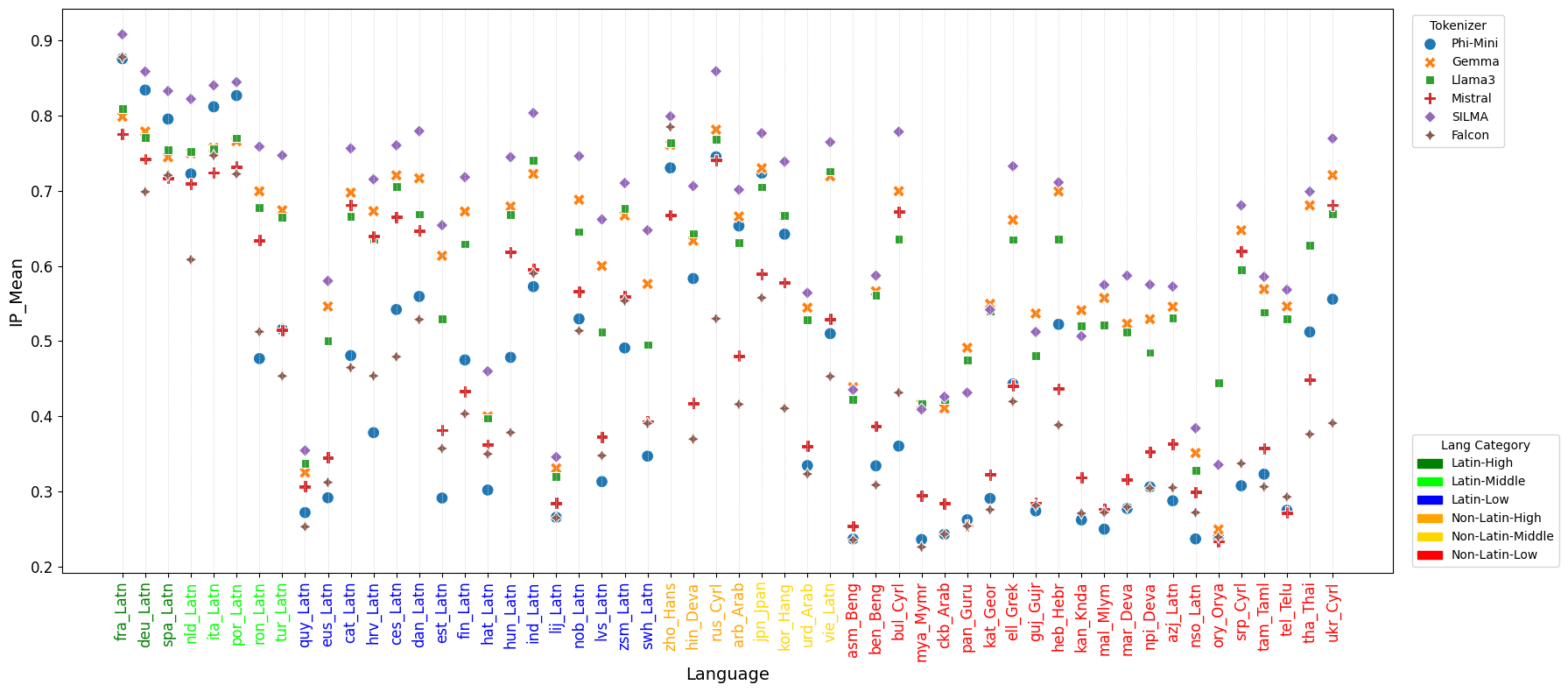}
    \caption{Information Parity (IP) across languages resulting from decoder-type models.}
    \label{fig:IP_LLM}
\end{figure*}
% Since the multi-label samples in these datasets were relatively few, we analyzed the actual multi-label (MuL) samples separately. 
%\vspace{-3.5mm}
\subsection{Bias Metrics Measurements}
We measure Tokenization Parity (TP) and Information Parity (IP) across six multilingual models: Phi-Mini-3.5, Gemma-7B, LLaMA-3.2 (3B), Mistral-7B, SILMA-9B, and Falcon-7B. Although SILMA is Arabic-focused, it builds on the multilingual Gemma architecture. The initial evaluation is conducted on 54 languages and dialectal variants from the FLORES-200 dataset—a parallel corpus of 2,000 human-translated Wikipedia sentences across 200 languages \cite{costa2022no}. A subset of these score is then used for the correlation analysis on the dialect NLP tasks.
% \begin{figure*}
%     \centering
%     \includegraphics[width=1\linewidth]{TP_Enc.png}
%     \caption{Tokenization Parity (TP) across Languages by Encoder-Model}
%     \label{fig:TP_Enc}
% \end{figure*}
% \begin{figure*}
%     \centering
%     \includegraphics[width=1.0\linewidth]{TP_Enc_nw.png}
%     \caption{Tokenization Parity (TP) across Languages by Encoder-Model}
%     \label{fig:TP_Enc_nw}
% \end{figure*}

% \begin{figure*}
%     \centering
%     \includegraphics[width=1.0\linewidth]{TP_LLM.png}
%     \caption{Tokenization Parity (TP) across Languages by Decoder-Model}
%     \label{fig:TP_LLM}
% \end{figure*}

\section{Results}
Table \ref{tab:Results} shows the F1-scores on the DI task. The encoder-type models outperform the heavily pre-trained decoder-type models across all datasets in both mono-label (ML) and multi-label (MuL) setups. Language-specific BERT models score better than mBERT in all cases except for the Swiss-German.\footnote{For curiosity, we tested also SILMA, the best performing Arabic decoder-type LLM on the NADI dataset. Although being an Arabic-specific model, it lags behind the mBERT model by almost 6 points and the Arabic-specific BERT model by about 28 points}. 
% \begin{figure*}
%     \centering
%     \includegraphics[width=1\linewidth]{IP_LLM.png}
%     \caption{Information Parity (IP) across Languages by Decoder-Model}
%     \label{fig:IP_LLM}
% \end{figure*}
% \begin{figure*}
%     \centering
%     \includegraphics[width=1\linewidth]{IP_enlarged.png}
%     \caption{Information Parity (IP) across languages resulting from decoder-type models.}
%     \label{fig:IP_LLM}
% \end{figure*}

\begin{figure*}
    \centering
    \includegraphics[width=1\linewidth]{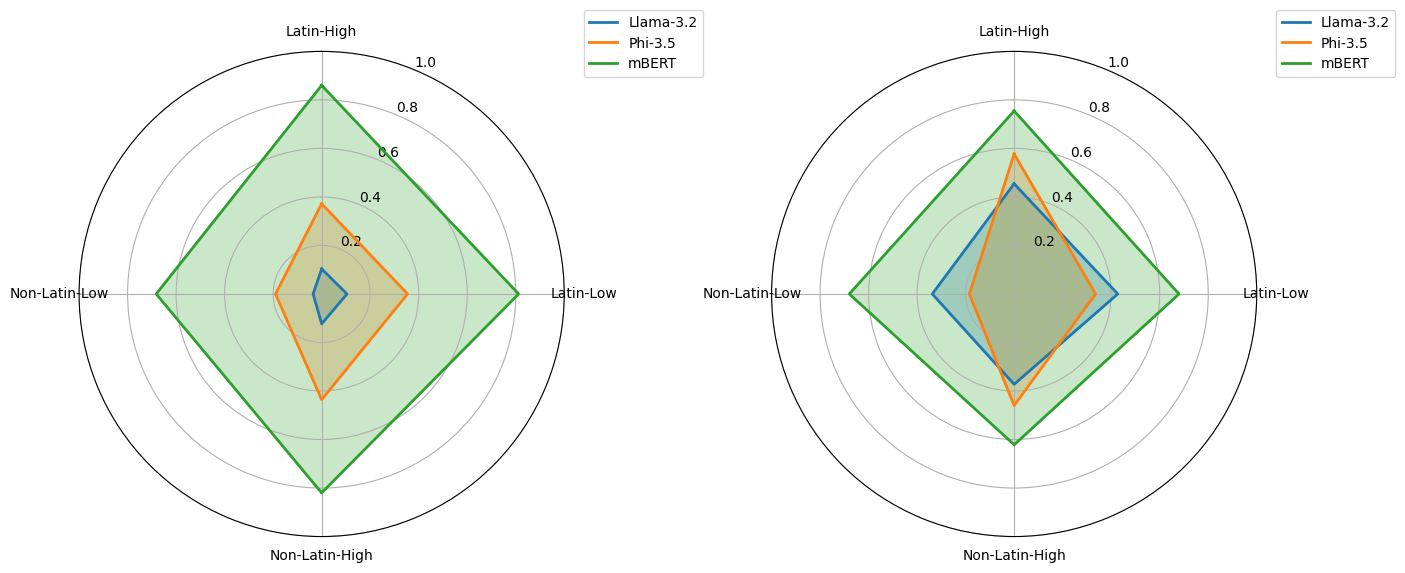}
    \caption{Average performance (F1-score) of models per category on TC (left) and EQA (right) tasks.}
    \label{fig:radar_avg}
\end{figure*}

% \begin{figure*}
%     \centering
%     \includegraphics[width=1\linewidth]{IP_enlarged.png}
%     \caption{Information Parity (IP) across Languages by Decoder-Model}
%     \label{fig:IP_LLM}
% \end{figure*}
% In the scores reported in Table \ref{tab:Results} as multi-label-only (MuLO), decoder-type models appear to perform better than mBERT, but this was an artifact of the evaluation method, which allows high scores if a model always outputs the same multilabels. 
%due to the fact that it consistently predicted a multi-label output (Discussed in Appendix \ref{app:perf}).
\begin{table*}
\centering
\scalebox{0.85}{
\begin{tabular}{p{3.9cm}p{1.5cm}p{2cm}p{1.5cm}p{5cm}}
\toprule
\textbf{Language (Type)} & \multicolumn{2}{c}{\textbf{Decoder-only}} & \multicolumn{2}{c}{\textbf{Encoder-only}} \\
\cmidrule(l){2-3}
\cmidrule(l){4-5}
& Phi-3.5 & Llama 3.2  & mBERT & Language-Specific \\
\midrule
Arabic (ML) & 0.54 & 0.26  & 0.62 & \textbf{0.84} (MARBERT) \\
Swiss-German (ML) & 0.49 & 0.46  & 0.59 &\textbf{0.60} (SwissBERT) \\
Indo-Aryan (ML) & 0.74 & 0.32  & 0.88 & \textbf{0.90} (IndicTransformers) \\
French (MuL) & 0.61 & 0.35 &  0.70 & \textbf{0.75} (CamemBERT) \\
Spanish (MuL) & 0.40 & 0.79 &  0.83 & \textbf{0.85} (spanBERTa) \\
% French (MuLO) & \textbf{0.61} & 0.35 & 0.28 & 0.44 (CamemBERT) \\
% Spanish (MuLO) & 0.53 & \textbf{1.00}  & 0.78 & 0.89 (spanBERTa) \\
\bottomrule
\end{tabular}}
\caption{\label{tab:Results} Performance (F1-scores) on dialect identification task across models. \textbf{ML} = Mono-label version of the task, \textbf{MuL} = Multi-label version of the task.}
% \textbf{MuLO} = Multi-label Only.}
\end{table*}
Figure \ref{fig:radar_avg} shows the results for the TC and EQA tasks. Here we report F1-score averages for the script and resource level groups, the detailed tabular results per dialectal variety are presented in Tables \ref{tab:phi_llama_f1_TC} and \ref{tab:grouped_f1_em} in Appendix \ref{app:results_details}.
On these tasks too, the encoder-type model, mBERT performs much better than the fine-tuned decoder multilingual LLMs. Regarding the controlled categories, it can be noted that the resource level affected the performance more than the script (the skewness of the polygons to the right), especially in decoder-type models. The differences between the model-types are smaller on the EQA task, as well as the impact of the resource level (except for Phi-3.5). Even though the impact of the script is smaller than that of the resource level, a bias towards Latin scripts is present, especially on the EQA task. 

\subsection{The distribution of the bias metrics values across languages}
Figures \ref{fig:TP_Enc_nw} and  \ref{fig:TP_LLM} show the distribution of the TP score on the sample of 54 FLORES languages sorted (and colored) according to the controlled categories (resource level and script type). A comparison of these two graphs shows that encoder-type tokenizers result in a more stable TP than the tokenizers of the decoder-type models. 
However, a clear divide emerges in both model types:  Latin-script languages maintain relatively stable TP and closer to 1 across all models, whereas non-Latin languages show substantial variability—particularly in lower-resource settings. Among decoder-type models, Gemma and SILMA demonstrate more consistent TP across language groups, while others show language-specific disparities.

When the TP values deviate more from 1, it shows larger disparities. For instance, with the mBERT tokenizer, the TP in German (Latin-High) is 1.26, while mBERT in Kannada (non-Latin-Low)is 2.19. This means the tokenizer produces 26\% more tokens for German than for English, which is a good tokenizer premium, indicating that German is fairly close to English in efficiency, since it uses Latin script and shares vocabulary with English. In contrast, with Kannada, the tokenizer produces 119\% more tokens than English for the same content, splitting the text into smaller fragments.

Figure \ref{fig:IP_LLM} illustrates IP performance (this score applies only to decoder-type models). High-resource Latin-script languages generally exhibit higher IP, while non-Latin and low-resource languages display wider variation. Unlike TP, IP appears to be more dependent on resource levels. 

% \begin{figure}
%     \centering
%     \includegraphics[width=0.5\linewidth]{heatmap_DI_nw.png}
%     \caption{Enter Caption}
%     \label{fig:enter-label}
% \end{figure}

% \begin{figure}
%     \centering
%     \includegraphics[width=0.5\linewidth]{heatmap_TC_nw.png}
%     \caption{Enter Caption}
%     \label{fig:enter-label}
% \end{figure}

% \begin{figure}
%     \centering
%     \includegraphics[width=0.5\linewidth]{heatmap_EQA_nw.png}
%     \caption{Enter Caption}
%     \label{fig:enter-label}
% \end{figure}
% \begin{figure*}
%     \centering
%     \begin{subfigure}[b]{0.32\textwidth}
%         \includegraphics[width=\textwidth]{heatmap_DI_nw.png}
%         \caption{Dialect Classification (DI)}
%         \label{fig:heatmap_di}
%     \end{subfigure}
%     \hfill
%     \begin{subfigure}[b]{0.32\textwidth}
%         \includegraphics[width=\textwidth]{heatmap_TC_nw.png}
%         \caption{Topic Classification (TC)}
%         \label{fig:heatmap_tc}
%     \end{subfigure}
%     \hfill
%     \begin{subfigure}[b]{0.32\textwidth}
%         \includegraphics[width=\textwidth]{heatmap_EQA_nw.png}
%         \caption{Extractive QA (EQA)}
%         \label{fig:heatmap_eqa}
%     \end{subfigure}
%     \caption{Correlation heatmaps showing Tokenization and Information Parity across dialectal tasks for different models. Blue indicates negative correlations and red positive correlations.}
%     \label{fig:correlation_heatmaps}

% \end{figure*}

\subsection{Correlation analysis of TP \& IP metrics with the downstream tasks}

To examine whether trends in Tokenization Parity (TP) and Information Parity (IP) across languages correlate with model performance on dialectal downstream tasks, we compute Pearson correlation coefficients between downstream task scores and the TP/IP metrics, using fine-tuned versions of Phi-3.5-Mini, LLaMA-3.2, and mBERT. Note that the direction of the correlation score is important for a meaningful interpretation of the results. 

In the case of TP, scores closer to 1 are considered better, while TP $>$ 1 indicates that the tokenizer uses more tokens to encode the same content compared to English (more fragmentation). Intuitively, we would expect a negative correlation between the value of TP and the downstream performance. High text fragmentation compared to the reference means that the input sequences are longer, which increases the complexity of the attention mechanism and makes modeling harder, which can impact the performance. In contrast, the expected correlation between IP and downstream performance is intuitively strongly positive, since a higher IP indicates greater representational efficiency, which should have a positive impact on the performance.
% In the case of TP, scores closer to 1 are better. Theoretically, a stronger positive correlation can be counterintuitive, while a stronger negative correlation could be what we expect. This is because, when TP $>$ 1, it means the tokenizer fragments the input heavily (as explained in Section \ref{sec:bias-metrics}, hence a strong positive correlation means when the tokenizer fragments more, the performance is better. Due to this reason, we present the directions of correlations and not just absolute correlation values for better interpretations. The expected correlation between the IP and downstream performance in general could be strongly positive. Since, in general, A higher IP value indicates greater representational efficiency.
% In the case of TP, lower scores are better, so we expect negative correlation with downstream performance, while the correlation should be positive between the IP and downstream performance. 

\begin{figure*}
    \centering
    \begin{subfigure}[b]{0.45\textwidth}
        \includegraphics[width=\textwidth]{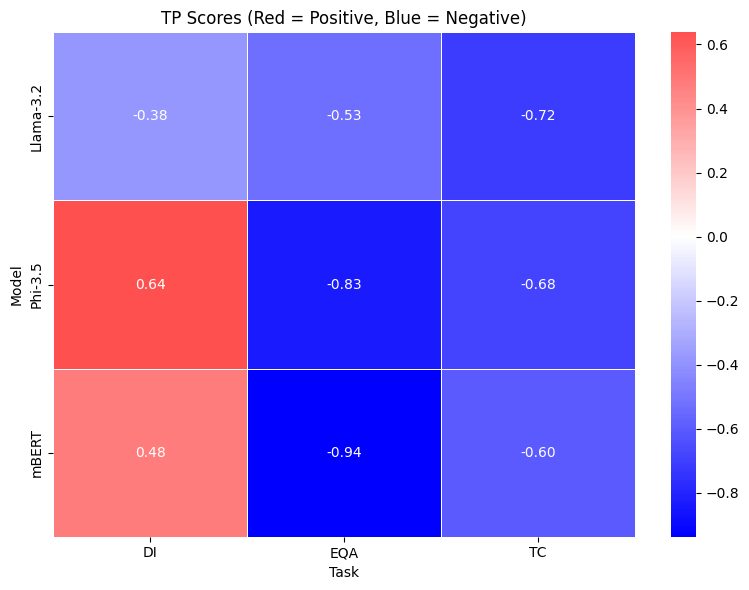}
        \caption{Correlations to TP across tasks and models}
        \label{fig:heatmap_tp}
    \end{subfigure}
    \hfill
    \begin{subfigure}[b]{0.45\textwidth}
        \includegraphics[width=\textwidth]{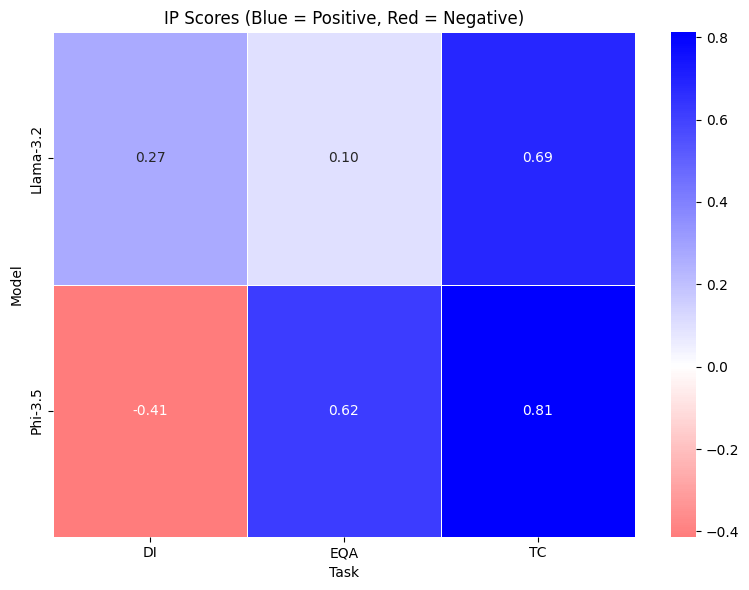}
        \caption{Correlations to IP cross tasks and models}
        \label{fig:heatmap_ip}
    \end{subfigure}
    % \hfill
    % \begin{subfigure}[b]{0.32\textwidth}
    %     \includegraphics[width=\textwidth]{heatmap_EQA_nw.png}
    %     \caption{Extractive QA (EQA)}
    %     \label{fig:heatmap_eqa}
    % \end{subfigure}
    \caption{Correlation heatmaps showing Tokenization and Information Parity across dialectal tasks for different models; \textcolor{blue}{blue} for the expected, \textcolor{red}{red} for the opposite of the expected.}
    \label{fig:correlation_heatmaps_nw}
\end{figure*}

Figure \ref{fig:correlation_heatmaps_nw} visualizes these correlations as heatmaps, with detailed tabular values provided in Appendix \ref{app:results_details}, Table \ref{tab:correlation_values}. To make the trends easier to follow, color codes show the expected correlations: blue for the expected, red for the opposite.  
%Also Note for encoder type mBERT there is no IP values
%The heatmaps also encode the direction of the correlation—positive (red) or negative (blue)—which is especially informative in interpreting TP.
\paragraph{Dialect Identification (DI)}
Contrary to the expected direction, we see a positive correlation between TP and DI performance in the two models that perform better (mBERT and Phi-3.5 in the map Figure \ref{fig:heatmap_tp}, cf. Table \ref{tab:Results}), while the expected negative correlation is observed only in Llama-3.2, whose performance is low. On the other hand, higher IP, reflecting more efficient information compression, is correlated with worse performance on dialect classification in Phi-3.5 (map Figure \ref{fig:heatmap_ip}). This outcome is also contrary to what we expected. The fact that the correlation is positive in Llama-3.2 only confirms this observation because the low performance of Llama-3.2 indicates that the task was not learned, and the model might be performing some other classification. Note also that the correlations are stronger in models that perform the task better. 
%Overall, TP appears to have stronger correlations across models, indicating that tokenization has a more significant impact on the DI task performance than ensuring even information representation. 

While we expected that higher tokenization disparity would lead to a performance drop, another picture appeared: it turns out that more fragmented text (compared to the reference), might, in fact, help models make surface-level distinctions if the task is learned at all. 
This could be attributed to the fact that dialects differ mostly at the surface level (spelling, morphology, and token patterns). If diacritics or other surface-level phenomena end up encoded as separate tokens due to higher text fragmentation, they might be exploited by models as useful dialect features even if the meaning of these units is not well captured in their vector representation. In other words, models do not need to ``understand'' the meaning of the small fragments to grasp their dialect specificity. In contrast, higher IP scores (expressing more equal compression) can be indicative of deeper level (semantic) similarity between the texts written in different dialects, making their differentiation harder even if the meaning is better captured. This would explain the surprising negative correlation between the IP score and the performance on the DI task.    
% IP seems less consistently impactful because dialectal variation is manifested as the surface-level differences (captured in tokens), rather than abstract information consistency. 
% From Figure \ref{fig:heatmap_di}, we observe a good correlation of TP to DI task for the Phi model. The positive correlation indicates that Phi-3.5's performance on DI improves as TP increases. This indicates that tokenization balance across languages significantly impacts Phi-3.5 in DI tasks. A negative correlation in IP suggests that higher IP is associated with lower performance for Phi-3.5, which could point to model biases or over-sensitivity to uniform information content. In Llama, a weak negative TP correlation implies that lower TP could help Llama-3.2's performance. A weak positive correlation with IP, indicating modest gains in performance with more balanced information access across languages. A moderately strong positive TP correlation suggests that mBERT performs better when TP is higher.
\vspace{-2mm}
\paragraph{Topic Classification (TC)} Comparing the two maps in Figure \ref{fig:correlation_heatmaps_nw}, we can see that IP is more strongly correlated with downstream performance than TP on this task, which applies more to the better model (Phi-3.5) than to the one with worse performance (Llama-3.2). This suggests that improved information compression across languages enhances performance on the TC task, but TP also shows a moderate correlation, indicating that tokenization may still impact the performance. 
\vspace{-1.75mm}
\paragraph{Extractive Question Answering (EQA)} It is especially interesting to see in Figure \ref{fig:correlation_heatmaps_nw}a that the correlation between TP and the performance on this task is strong both in mBERT and Phi-3.5. This suggests that variation in tokenization can significantly impact the model's ability to extract the correct span. There is also a moderate correlation with IP (the map Figure \ref{fig:heatmap_ip}), indicating that more consistent information representation across dialects may help the model extract relevant answers more effectively. In contrast, Llama-3.2 shows a moderate correlation with TP, but the correlation with IP is negligible. These findings suggest that tokenization disparities play a more significant role than general information preservation in extractive QA tasks, where accurate token-level span prediction is crucial.
% \bigskip
% \noindent
%Overall, IP generally correlates well with semantic tasks such as topic classification, while the effect are less consistent in the case of dialect identification. This finding suggests that high tokenization TP does not necessarily lead to Since IP measures the efficiency of information representation, it assumes that a language can encode the same content as English with similar compression. 

%However, as discussed, dialectal differences are often expressed at the surface level, through orthography, diacritics, or morphological markers, rather than in high-level semantic content. High IP may smooth over these surface-level cues, effectively making dialects appear more similar and reducing task performance. In contrast, TP captures token-level fragmentation, which preserves these subtle distinctions, sometimes even facilitating dialect identification. This indicates that IP is most informative when semantic consistency drives task performance, but less so when surface-level variation is the main signal, as in dialect identification.

Taken together, our results suggest that higher TP (more tokens per word for some languages) usually hurts the performance in both surface-level and semantically rich tasks as long as semantic representations are needed for the task. Higher IP scores (more similar compression), on the other hand, are associated with better downstream performance in both cases. The stronger association between the TP scores and the surface-level tasks, on one hand, and between the IP scores and the semantically rich tasks, on the other, is in line with the previous results reported by \citet{tsvetkov2024information}, where TP was better correlated with extractive or text similarity tasks (e.g., PAWSX, XQuAD). At the same time, IP correlated better with tasks requiring semantic consistency (e.g., reasoning), corresponding to our TC setting. The results on the DI task suggest an inverse relation between the TP and IP scores and the downstream task, which has not been reported in previous studies. In this case, higher TP is associated with better performance, while higher IP with worse downstream performance. As discussed above, the explanation for these effects comes from the fact that the task of distinguishing between dialect does not require deep semantic representation of surface-level features, while deeper semantic similarity (potentially captured by a high IP score) can even hurt the performance making the dialects harder to distinguish.  

\subsection{A closer look at the Llama tokenizer}
\begin{figure}
\bigskip
\noindent
\textbf{English reference}: \\
\emph{The find also grants insight into the evolution of feathers in birds}

% \bigskip
\noindent
\textbf{Arabic}: \\
\selectlanguage{arabic}{يمنح الاكتشاف أيضاً نظرة على تطور الريش في الطيور}
\selectlanguage{english}

% \bigskip
\begin{verbatim}
Llama3_Tokenizer_Output: ['ÙĬ', 'ÙħÙĨ', 
'ØŃ', 'ĠØ§ÙĦ', 'Ø§Ùĥ', 'ØªØ´', 'Ø§Ùģ', 
'ĠØ£ÙĬØ¶Ø§', 'Ùĭ', 'ĠÙĨØ¸', 'Ø±Ø©', 
'ĠØ¹ÙĦÙī', 'ĠØª', 'Ø·ÙĪØ±', 'ĠØ§ÙĦ', 
'Ø±ÙĬ', 'Ø´', 'ĠÙģÙĬ', 'ĠØ§ÙĦØ·', 
'ÙĬ', 'ÙĪØ±']
\end{verbatim}
\caption{Example of Llama-3.2 tokenizer output.} 
\label{fig:Llama_Ar}
\end{figure}
% \vspace{-5mm}
As the Llama-3.2 model performance was behind all other models on all tasks, we take a closer look into its tokenizer and how it deals with non-Latin scripts. For this, we output the tokenization of a given input text as shown in the Arabic example in  Figure \ref{fig:Llama_Ar}.
It can be observed that Llama-3.2 outputs misaligned tokens, which turned out to be characters misinterpreted as Latin-1, which is induced by the byte-level fallbacks. For instance, the token \texttt{ Ø³ } has the following Unicode description: ['LATIN CAPITAL LETTER O WITH STROKE,' 'SUPERSCRIPT THREE']. The characters may break into smaller byte-level components if not directly present in the tokenizer vocabulary. These byte sequences may be aligned to Latin-character tokens due to a high bias toward Latin script in pre-training data. During decoding, the tokenizer may reassemble these tokens into the correct Unicode characters that match the non-Latin script. However, this can degrade the performance of non-Latin language tasks, as the model may not be able to capture the semantics and produce longer token sequences. Also, this script raises questions about how well the model captures the semantic meaning and linguistic nuances. 

The same behavior was also noted in the non-Latin script language Hindi. For instance, {\dn rAvn} was tokenized as ['à¤°', 'à¤¾à¤µà¤¨'], where the Hindi character {\dn{r}} corresponds to the UTF-8 bytes [E0 A4 B0] \footnote{\url{https://www.utf8-chartable.de/unicode-utf8-table.pl?start=2304&number=128}}. This sequence is interpreted as [à ¤ °] in Latin-1, which is then represented as the token \texttt{ à¤° }. Similar observations were made in all non-Latin scripts experimented with, where Latin characters were recognized. It should be noted, though (from Appendix \ref{app:tok_details}) that both Phi-3 and Llama-3 tokenizers are based on TikToken. This means that the tokenization behavior also largely depends on the tokenizer's knowledge of the pre-trained language.

As additional analyses, we examined the correlation between missing character proportions and downstream performance and investigated the language support specifications of the LLMs. Our findings suggest that these relationships remain highly model- and task-dependent. Detailed results and discussions are provided in Appendix \ref{app:analysis}.

\section{Related Work}\label{app:rel}
% The importance of tokenization in LLMs, language fairness, and the proposal of multilingually fair tokenizers are some of the main lines of research in this direction. 
The fairness and biases of LLM tokenizations have been analyzed using parallel language corpora by \citep{petrov2024language,ahia2023all,  rust2021good}. Language-specific investigations \cite{toraman2023impact}, temporal evaluations \cite{spathis2024first}, and adversarial impacts \cite{wang2024tokenization} and tokenizer comparisons \cite{kanjirangat2023optimizing,batsuren2024evaluating} are other directions. The general conclusion pinpointed the importance of tokenization - \emph{tokenization matters}. Following the limitations of tokenizers and other multilingual biases, another research dimension proposes alternative tokenization approaches \cite{hofmann2022embarrassingly} and even tokenless models \cite{barrault2024large,pagnoni2024byte}. Extending the understanding and analysis of representational biases in multilingual LLMs, some potential works on metrics related to information theory perspectives \cite{tsvetkov2024information} and \cite{land2024fishing}.
The primary line of existing research in dialectal tasks focus on performance improvements across various datasets using LLMs \cite{scherrer2024proceedings,alam2024llms,frei2023automatic}, with a recent focus on multi-label DI \cite{bernier2023dialect,keleg2023arabic,chifu2024vardial,kanjirangat2024nlp_di}. The primary research focused on assessing GPT-based models' multilingual capabilities, highlighting their limitations, with a few exceptions. GPT capability in Arabic was evaluated in \cite{khondaker2023gptaraeval}, unveiling the shortcomings of dialectal Arabic and the supremacy of encoder models. In \cite{lai2023chatgpt,bang2023multitask}, ChatGPT was evaluated in diverse languages, showing the predominance of high-resource languages. Recently, a comprehensive dialectal benchmark dataset was introduced, DialectBench \cite{faisal2021sd}, which encompasses various dialectal tasks covering a wide range of dialectal varieties. While there has been notable research in dialectal tasks and multilingual NLP individually, efforts to bridge the two remain limited. Existing work has largely focused on performance comparisons, with less attention to understanding the underlying causes of degraded performance.

\section{Conclusion}\label{sec:con}
%\vspace{-5mm}
In this paper, we go beyond traditional performance-based evaluations of dialectal downstream tasks to examine multilingual fairness and potential biases arising from disparities in tokenization and representation. 
%language scripts and resource levels. 
We show that  
%two quantitative metrics—
Tokenization Parity (TP) and Information Parity (IP) 
%—
%and analyze their 
correlate with downstream task performance in a consistent, although sometimes surprising, way. Our results reveal consistent disparities in TP between Latin and non-Latin scripts, while IP variations are influenced by both script and resource availability. TP is more strongly associated with tasks involving syntactic, morphological, and span-based features, whereas IP aligns more closely with tasks requiring semantic understanding and reasoning. The role of token-level disparity is especially interesting in surface-level tasks such as dialect identification, which can help models make distinctions between dialects. As a future direction, we emphasize the importance of developing language-aware, adaptive tokenizers that can mitigate pre-training biases and flexibly operate across multiple levels of granularity. 
%Another potentially valuable research direction lies in the development of fairness-aware evaluation suites.
% This paper provides an in-depth analysis of tokenization and language-related factors in pre-trained LLMs, aiming to enhance our understanding of their multilingual capabilities and possible influences on the downstream use cases. Key aspects such as parity, language support, and vocabulary coverage were examined to assess the fairness of these models across Latin and non-Latin languages. A persistent script bias favoring Latin-based languages was observed, ultimately affecting the equitable performance of these models in multilingual contexts. To assess the influence of these factors on language-dependent downstream tasks, we explored dialect classification as a use case. Our findings aligned with earlier observations, revealing performance biases in LLMs across languages and notable disparities compared to encoder variants. Additionally, language-specific models demonstrated consistent superiority in most cases, raising questions about the applicability of multilingual models for monolingual tasks, particularly in non-Latin script scenarios. These insights underline the need for more equitable tokenization strategies and broader language support in future LLM development. We propose improved tokenization strategies and language representations as future recommendations, detailed in Appendix \ref{app:recommend} (due to space constraints).

\section{Limitations}
There is significant scope for further enhancing the LLMs examined in this work. The tokenization analysis can be improved by leveraging more extensive and diverse corpora, enabling more profound insights into tokenization strategies and their implications. While the primary focus here was to analyze the relationship between tokenization, language-specific factors, and their impact on language-dependent tasks, future work could explore additional use cases by identifying and incorporating relevant tasks. In this study, we concentrated on the dialectal tasks: first, as a challenging language-dependent task, it offers a robust testbed for examining tokenization impacts; and second, this aspect has largely been overlooked in prior research, where the emphasis has predominantly been on performance metrics. Expanding this investigation to include other complex language-dependent tasks could further elucidate the role of tokenization in multilingual LLM performance. 
\section*{Acknowledgments}
This work was supported by the project "fairTOK", funded by armasuisse S\&T. 
The authors also gratefully acknowledge the reviewers for their insightful and constructive feedback, which helped improve the quality of this work.
\bibliography{custom}
%\clearpage 
\clearpage
\appendix
\input{appendix.tex}
\end{document}

%% file: appendix.tex
\appendix
\renewcommand\thesubsection{\thesection.\arabic{subsection}}
\begin{figure}
    \centering
    \includegraphics[width=1\linewidth]{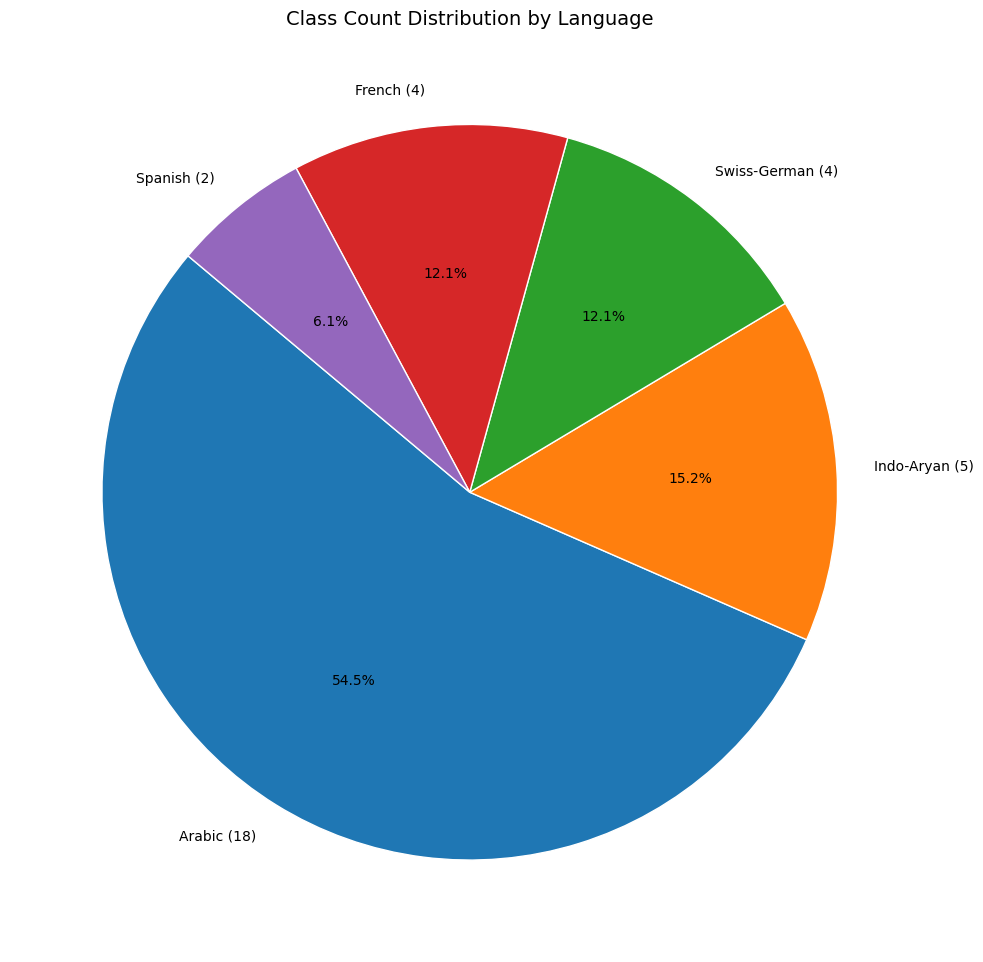}
    \caption{Class distributions of dialect classification task (Appendix \ref{app:data})}
    \label{fig:Class_distribution_DI}
\end{figure}
 \begin{table*}[h!]
\centering
\renewcommand{\arraystretch}{0.5}
\begin{tabular}{lrrrrr}
\hline
{} & \textbf{GDI} & \textbf{ILI} & \textbf{NADI} & \textbf{DSL-ML-FR} & \textbf{DSL-ML-ES} \\ 

\hline
Train & 14647 &  68453 & 18000 & 340363 & 3467  \\
Test  & 4752 &  9032 & 1800 & 17090 & 989 \\ 
%Test & 4752 & 9032 & 1492 & 10812 & 4871\\
No. of labels  & 4 & 5 & 18 & 4 & 2\\

\hline
\end{tabular}

\caption{\label{tab:datas} Dataset statistics (Appendix \ref{app:data})}
\end{table*}

\section{Dataset Details}\label{app:data}
\begin{figure}
    \centering
    \includegraphics[width=1\linewidth]{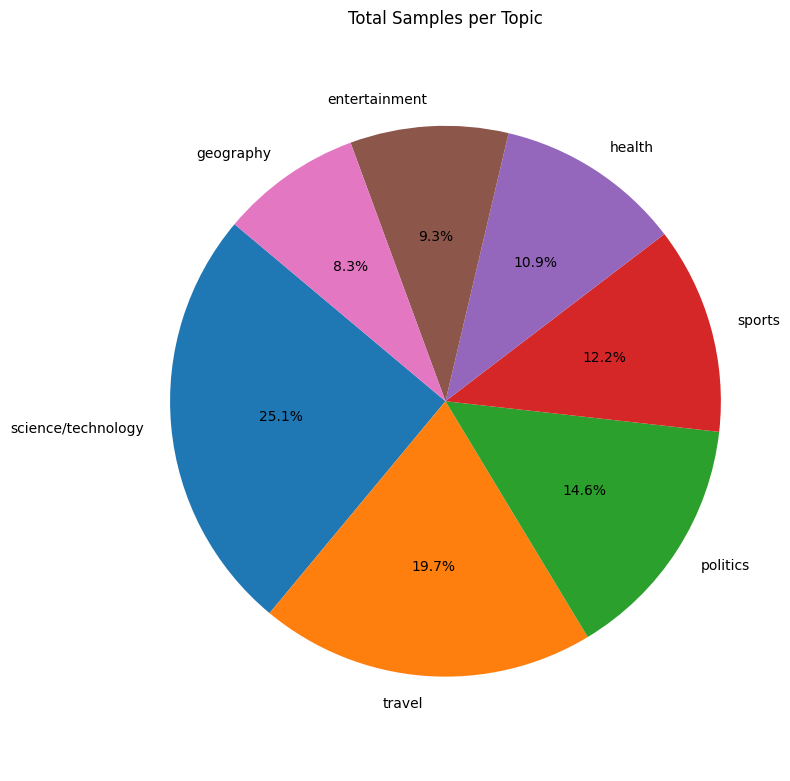}
    \caption{Class distributions topic classification task (Appendix \ref{app:data})}
    \label{fig:Class_distribution_TC}
\end{figure}

\begin{table*}
\centering
\renewcommand{\arraystretch}{0.5}
\begin{tabular}{lp{0.5\textwidth}l}
%{lllrrrrr}

\hline
\textbf{Models} & \textbf{Tokenizer}  & \textbf{Model\_Vocab\_Sizes}\\
\hline
Phi3  &  tiktoken \cite{abdin2024phi} \footnote{\url{https://github.com/openai/tiktoken}} & 32011\\
Gemma  & SentencePiece tokenizer\cite{kudo2018sentencepiece,team2024gemma} & 256000\\
Llama3  & tiktoken \cite{dubey2024llama} & 128256\\
Bloom & Byte-level BPE \cite{wang2020neural} & 250680\\
Mistral  & tekken- Modified tiktoken \cite{jiang2024mixtral} & 32000\\ 
%Meltemi & tekken (Based on Mistral-7B) & 61377\\
NLLB &SentencePiece tokenizer tailored &256204\\
BERT-based & WordPiece \cite{song2021fast,devlin2018bert}& 30522 \\
MARBERT & WordPiece &100000\\
mBERT & WordPiece &119547\\
IndicBERT & WordPiece &200000\\
SpanBERTa & WordPiece &50265\\
CamemBERT & WordPiece & 32005\\
%GreekBERT & WordPiece & 35000\\
 \hline
\end{tabular}
\caption{\label{tab:tok} Tokenizers and vocabulary Sizes of LLMs (Appendix \ref{app:tok_details})}
\end{table*}

\begin{table*}[ht]
\centering
\footnotesize
\renewcommand{\arraystretch}{1.2}
\begin{tabularx}{\linewidth}{lX}
\toprule
\textbf{Category} & \textbf{Languages (Flores Code)} \\
\midrule
Latin-High &
Spanish (spa\_Latn), German (deu\_Latn), French (fra\_Latn) \\
Latin-Middle &
Dutch (nld\_Latn), Italian (ita\_Latn), Romanian (ron\_Latn), Turkish (tur\_Latn), Portuguese (por\_Latn) \\
Latin-Low &
Ayacucho Quechua (quy\_Latn), Haitian Creole (hat\_Latn), Basque (eus\_Latn), Hungarian (hun\_Latn), Catalan (cat\_Latn), Danish (dan\_Latn), Estonian (est\_Latn), Indonesian (ind\_Latn), Standard Latvian (lvs\_Latn), Standard Malay (zsm\_Latn), Finnish (fin\_Latn), Swahili (swh\_Latn), Norwegian Bokmål (nob\_Latn), Croatian (hrv\_Latn), Czech (ces\_Latn), Ligurian (lij\_Latn) \\
Non-Latin-High &
Standard Arabic (arb\_Arab), Russian (rus\_Cyrl), Chinese (Simplified) (zho\_Hans), Hindi (hin\_Deva) \\
Non-Latin-Middle &
Urdu (urd\_Arab), Korean (kor\_Hang), Vietnamese (vie\_Latn), Japanese (jpn\_Jpan) \\
Non-Latin-Low &
North Azerbaijani (azj\_Latn), Thai (tha\_Thai), Marathi (mar\_Deva), Odia (ory\_Orya), Gujarati (guj\_Gujr), Nepali (npi\_Deva), Burmese (mya\_Mymr), Assamese (asm\_Beng), Central Kurdish (ckb\_Arab), Tamil (tam\_Taml), Malayalam (mal\_Mlym), Bulgarian (bul\_Cyrl), Eastern Panjabi (pan\_Guru), Ukrainian (ukr\_Cyrl), Bengali (ben\_Beng), Kannada (kan\_Knda), Greek (ell\_Grek), Northern Sotho (nso\_Latn), Serbian (srp\_Cyrl), Telugu (tel\_Telu), Hebrew (heb\_Hebr), Georgian (kat\_Geor) \\
\bottomrule
\end{tabularx}
\caption{Language categories and corresponding languages with FLORES codes: Appendix \ref{app:category_details} }
\label{tab:language_groups}
\end{table*}

Table \ref{tab:datas} shows the general statistics of DI task datasets. In the DI task, NADI-2023 has 18 dialects from Arabic-speaking regions such as Iraq, Oman, Saudi Arabia, Palestine, Bahrain, Egypt, Jordan, Libya, Sudan, UAE, Algeria, Kuwait, Tunisia, Lebanon, Morocco, Yemen, Syria, and Qatar. In GDI, we had four dialects: Zurich, Luzern, Basel, and Bern. For ILI, it was Hindi, Braj Bhasha, Awadhi, Bhojpuri, and Magahi. In multi-label settings, we used the datasets from the Multi-label classification of similar languages (DSL-ML) 2024 shared task, focusing on manually labeled Spanish and automatically labeled French data. For French, data was from the FreCDo dataset, including French (FR-FR), Swiss (FR-CH), Canadian (FR-CA), and Belgian (FR-BE) with  \{'FR-BE': 120653, 'FR-CH': 115664, 'FR-FR': 83127, 'FR-CA': 19041, 'FR-BE, FR-FR': 1052, 'FR-BE, FR-CH': 603, 'FR-CH, FR-FR': 162, 'FR-BE, FR-CH, FR-FR': 61\} as multi-label samples. For Spanish, the two varieties were Argentinian and peninsular Spanish, with 1131 multi-label samples.

Under the multi-lingual setup, we created a representative train-test sample dataset for the French dataset due to the massive size of the automatically curated dataset. We selected 5000 mono-label samples from each class and all the multi-label samples comprising the train set. 1000 mono-label samples with all multi-label samples were selected for the test set. This constitutes 21878 (20000 (mono)+ 1878 (multi)) train and  4120 (4000+120) test samples. 

The class distributions of the DI and TC tasks are shown in Figures \ref{fig:Class_distribution_DI} and \ref{fig:Class_distribution_TC}.
%%%%%%%%%%%%%%%%%%%%%%%%%%%%%%%%%%%%%%%%%%%%%%%%%%%%%%%%%%%%
\section{Tokenizer details}\label{app:tok_details}
The details of the tokenizers and pre-trained vocabulary sizes used by the evaluated models are shown in Table \ref{tab:tok}.
%%%%%%%%%%%%%%%%%%%%%%%%%%%%%%%%%%%%%%%%%%%%%%%%%%%%%%%%%%%%%%%%%%%%%%%%%%%%%%%%%%%%%%
%\section{Method details}\label{app:method_details}
\section{Script \& resource Categorizations}\label{app:category_details}
The details of the languages under different scripts and resource categories are shown in Table \ref{tab:language_groups}.
\section{Prompt details}\label{app:prompts}
This section presents the instruction-tuning prompts used for the experiments in decoder-only LLMs. Figures \ref{fig:prompt1}, \ref{fig:prompt2}, and \ref{fig:prompt3} represent the prompts for DI, TC, and EQA tasks, respectively.
%%%%%%%%%%%%%%%%%%%%%%%%%%%%%%%%%%%%%%%%%%%%%%%%%%%%%%%%%%%%%%%%%%%%%%%%%%%%%%%%%%%%%%
\section{Experimental settings details}\label{app:exp}
We evaluated the experiments on HPC clusters with a100 and v100 GPUs. The runtime varied between approximately. 4 hours - 1.5 days for decoder models and 1-2 hours for encoder models. All experimental models were accessed from the HuggingFace library. 
\floatstyle{boxed}
\restylefloat{figure}
\begin{figure*}[htp!]
\centering
%\tiny
\begin{minipage}{0.9\textwidth}
\begin{verbatim}
TRAINING_CLASSIFIER_PROMPT = """[INST]What is the dialect of the given input 
sentence.
Sentence:{sentence} 
Class:{label}[/INST]"""
INFERENCE_CLASSIFIER_PROMPT = """[INST] Classify the dialect of the sentence. 
Choose from one of the following options:{allowed_labels}.
Sentence:
{sentence}
[/INST]
Class:"""
\end{verbatim}
\end{minipage}
\caption{Instruction-tuning prompt for dialect classification task (Appendix \ref{app:prompts})}\label{fig:prompt1}
\end{figure*}

\begin{table}
\centering
\renewcommand{\arraystretch}{0.5}
\begin{tabular}{p{2cm}p{5cm}}
%{lllrrrrr}
\hline
\textbf{Language} & \textbf{Unicode ranges \& special characters}  \\
\hline
Hindi  &(0x0900, 0x097F + 1)\\ 
Arabic &  (0x0600, 0x06FF + 1) \\
%Greek  & (0x0391, 0x03E1 + 1)\\
English &  (0x41, 0x5B) and (0x61, 0x7B)\\
% (65, 91) and (97, 123)
Spanish & ['á', 'é', 'í', 'ó', 'ú', 'ü', 'ñ', \\
&  'Á', 'É', 'Í', 'Ó', 'Ú', 'Ü', 'Ñ'] \\
French & ['à', 'â', 'ä', 'ç', 'é', 'è', 'ê', 'ë', \\ &'î','ï', 'ô', 'ö', 'ù', 'û', \\
        &'ü','À', 'Â', 'Ä', 'Ç', 'É', 'È', 'Ê',\\
        &'Ë', 'Î', 'Ï', 'Ô', 'Ö', 'Ù',\\
        &'Û', 'Ü']\\
Swiss-German &['ä', 'ö', 'ü', 'Ä', 'Ö', 'Ü', 'ß'] \\
 \hline
\end{tabular}
\caption{\label{tab:uni} Character Unicode ranges and special characters for different languages - all special characters are contained in the Latin-1 supplement Unicode block 0x0080-0x00FF. (Appendix \ref{app:vocab_uni})}
\end{table}
%%%%%%%%%%%%%%%%%%%%%%%%%%%%%%%%%%%%%%%%%%%%%%%%%%%%%%%%%%%%%%%%%%%%%%%%%%%%%%%%%%%%%%
\section{Result details}\label{app:results_details}
In this section, we present the detailed tabular results for the TC and EQA tasks per dialectal variety. Table \ref{tab:phi_llama_f1_TC} and \ref{tab:grouped_f1_em} present the results on TC and EQA tasks, respectively. Table \ref{tab:correlation_values} presents the detailed correlation values of TP and IP over different downstream tasks across Phi-3.5, Llama-3.2 and mBERT models. 
\begin{table*}[htb]
\centering
\renewcommand{\arraystretch}{1.1}
\scalebox{0.85}{
\begin{tabular}{llccc}
\toprule
\textbf{Category} & \textbf{Language} & \textbf{LLaMA-3.2} & \textbf{Phi-3.5 } & \textbf{mBERT}\\
\midrule
\multirow{8}{*}{Latin-High} 
& Dutch (nld\_Latn) & 0.1096 & 0.4178 & 0.894\\
& English (eng\_Latn) & 0.0494 & 0.3210&0.897 \\
& French (fra\_Latn) & 0.1436 & 0.4274 &0.910\\
& German (deu\_Latn) & 0.1187 & 0.3692& 0.862\\
& Italian (ita\_Latn) & 0.1137 & 0.3309& 0.872\\
& Portuguese (por\_Latn) & 0.0823 & 0.3523 &0.868\\
& Spanish (spa\_Latn) & 0.1313 & 0.3813& 0.821\\
& Romanian (ron\_Latn) & 0.0791 & 0.3762 &0.857\\

\midrule
\multirow{9}{*}{Latin-Low} 
& Catalan (cat\_Latn) & 0.1010 & 0.3502 &0.858\\
& Croatian (hrv\_Latn) & 0.0972 & 0.4823 &0.858\\
& Estonian (est\_Latn) & 0.1686 & 0.4508 &0.766\\
& Finnish (fin\_Latn) & 0.1241 & 0.3067 &0.809\\
& Haitian Creole (hat\_Latn) & 0.0700 & 0.1946 &0.61\\
& Hungarian (hun\_Latn) & 0.1206 & 0.3671 &0.861\\
& Indonesian (ind\_Latn) & 0.0937 & 0.3666& 0.847\\
& Norwegian Bokmål (nob\_Latn) & 0.1024 & 0.3592 &0.862\\
& Basque (eus\_Latn) & 0.0630 & 0.2946 &0.82\\

\midrule
\multirow{5}{*}{Non-Latin-High} 
& Arabic (arb\_Arab) & 0.1491 & 0.5194 &0.811\\
& Hebrew (heb\_Hebr) & 0.1507 & 0.3685 &0.83\\
& Hindi (hin\_Deva) & 0.0888 & 0.4972 &0.742\\
& Japanese (jpn\_Jpan) & 0.0704 & 0.4043 &0.888\\
& Russian (rus\_Cyrl) & 0.1565 & 0.3897 &0.827\\

\midrule
\multirow{7}{*}{Non-Latin-Low} 
& Bengali (ben\_Beng) & 0.0288 & 0.2192 &0.773\\
& Gujarati (guj\_Gujr) & 0.0000 & 0.0824 &0.597\\
& Kannada (kan\_Knda) & 0.0287 & 0.0952 &0.78\\
& Malayalam (mal\_Mlym) & 0.0089 & 0.0938 &0.66\\
& Marathi (mar\_Deva) & 0.1011 & 0.3732 &0.744\\
& Nepali (npi\_Deva) & 0.0503 & 0.3610 &0.762\\
& Orya (ory\_Orya) & 0.0281 & 0.1056 &0.461\\

\bottomrule
\end{tabular}}
\caption{\label{tab:phi_llama_f1_TC} Macro F1 scores for languages under different resource-script categories in the topic classification task. (Appendix \ref{app:results_details} )}
\end{table*}
% \begin{table*}[ht]
% \centering
% \begin{tabular}{lcc}
% \hline
% \textbf{Group} & \textbf{avg-F1} & \textbf{avg-EM} \\
% \hline
% \multicolumn{3}{c}{\textbf{Phi-3.5}} \\
% \hline
% Latin + High Resource     & 0.582109 & 0.353926 \\
% Latin + Low Resource      &0.335537 & 0.134534\\
% Non-Latin + High Resource & 0.459883 & 0.306647\\
% Non-Latin + Low Resource  & 0.184555 & 0.061947 \\
% \hline
% \multicolumn{3}{c}{\textbf{Llama-3.2}} \\
% \hline
% Latin + High Resource     & 0.455691 & 0.291322 \\
% Latin + Low Resource      & 0.427162& 0.324153\\
% Non-Latin + High Resource & 0.372376 &0.245423\\
% Non-Latin + Low Resource  &0.337537  &0.203540  \\
% \hline
% \multicolumn{3}{c}{\textbf{mBERT}} \\
% \hline
% Latin + High Resource     &  & \\
% Latin + Low Resource      & & \\
% Non-Latin + High Resource &  &\\
% Non-Latin + Low Resource  &  & \\
% \hline
% \end{tabular}
% \caption{Average Macro-F1 and Micro-F1 scores for Phi-3.5 and Llama-3.2 across four groups in Dialectal\_EQA.} 
% % \textbf{Latin + High Resource}: english; \textbf{Latin + Low Resource}: swahili,; \textbf{Non-Latin + High Resource}: Arabic, Korean; \textbf{Non-Latin + Low Resource}: Bengali.}
% \label{tab:phi_llama_EQA_avg}
% \end{table*}
\begin{table*}[ht]
\centering
\scalebox{0.85}{
\begin{tabular}{llccc}
\hline
\textbf{Category} & \textbf{Language Code} & \textbf{Llama3} & \textbf{Phi-3.5} &\textbf{mBERT} \\
\hline
\multirow{10}{*}{Latin-High} & english-kenya &0.431337  & 0.540717 & 0.725\\
 & english-nzl & 0.462533 & 0.596464 & 0.767 \\
 & english-irl & 0.462224 & 0.600408& 0.755\\
 & english-ind\_n & 0.443038 &0.573698 &0.746 \\
 & english-phl & 0.455140 & 0.588592& 0.764\\
 & english-nga & 0.452876 & 0.566533&0.736 \\
 & english-aus & 0.458323 & 0.593847&0.757 \\
 & english-ind\_s & 0.431355 & 0.543300& 0.719\\
 & english-usa & 0.479512 & 0.608536&0.772 \\
 & english-gbr & 0.471987 & 0.596156& 0.764\\
 & english-zaf & 0.464280 & 0.594952& 0.766\\
\hline
\multirow{2}{*}{Latin-Low} & swahili-kenya & 0.443695 &0.350143 &0.724 \\
 & swahili-tanzania & 0.410629 &0.320931 & 0.635\\
\hline
\multirow{9}{*}{Non-Latin-High} & arabic-sau & 0.361613 &0.457623 & 0.778\\
 & arabic-mar & 0.361082 &0.445782 & 0.767\\
 & arabic-jor & 0.360686 & 0.45141&0.773\\
 & arabic-tun & 0.358351 &0.448329 & 0.767\\
 & arabic-bhr & 0.359525 & 0.456000 &0.775 \\
 & arabic-dza & 0.357209 & 0.455835&0.778\\
 & arabic-egy & 0.345871 & 0.441766& 0.765\\
 & korean-korn & 0.432525 & 0.481986& 0.10\\
 & korean-kors & 0.414520 &0.500209 & 0.092\\
\hline
\multirow{2}{*}{Non-Latin-Low} & bengali-ind & 0.325579 & 0.176780&0.686 \\
 & bengali-dhaka & 0.349494 &0.192330 &0.673 \\
\hline
\end{tabular}
}
\caption{F1 and EM scores by language code and category for EQA task -Appendix \ref{app:results_details}}
\label{tab:grouped_f1_em}
\end{table*}

\begin{table*}[htb]
\centering
\renewcommand{\arraystretch}{1.3}
\scalebox{0.85}{
% \begin{tabular}{llcccc}
\begin{tabularx}{\textwidth}{l ccc}
\hline
\textbf{Task} & \textbf{Model / Category} & \textbf{Tokenization Parity} & \textbf{Information Parity} \\
\hline
\multirow{3}{*}{\textbf{Dialect Classification (DI)}} 
& Phi-3.5 & 0.638 & -0.413 \\

\cline{2-4}
& Llama-3.2 & -0.380 & 0.268 \\

\cline{2-4}
& mBERT & 0.4836 &\textemdash \\

\hline

\multirow{7}{*}{\textbf{Topic Classification (TC)}}
& Phi-3.5 & -0.683 & 0.812 \\
& \quad Latin-High & 0.873 &  0.765 \\
& \quad Latin-Low & -0.785 & 0.165 \\
& \quad Non-Latin-High & 0.202 & -0.862 \\
& \quad Non-Latin-Low & 0.876 & 0.634 \\
\cline{2-4}
& Llama-3.2 & -0.716 & 0.687 \\
& \quad Latin-High &  0.974& 0.671 \\
& \quad Latin-Low &   0.077& 0.328 \\
& \quad Non-Latin-High &   -0.547& 0.209 \\
& \quad Non-Latin-Low &   -0.805& 0.623 \\
\cline{2-4}
& mBERT & -0.605 & \textemdash \\
& \quad Latin-High &-0.242  &  \textemdash\\
& \quad Latin-Low & -0.706 & \textemdash \\
& \quad Non-Latin-High &-0.826  & \textemdash\\
& \quad Non-Latin-Low & -0.828 & \textemdash \\
\hline

\multirow{3}{*}{\textbf{Dialectal Extractive QA (EQA)}} 
& Phi-3.5 & -0.834 & 0.618 \\

\cline{2-4}
& Llama-3.2 & -0.528 & 0.097 \\

\cline{2-4}
& mBERT & -0.938 & \textemdash\\

\hline
% \end{tabular}
\end{tabularx}
}
\caption{Correlation values (overall and per category) between model tokenization/information parity and dialectal task performance-Appendix \ref{app:results_details}}
\label{tab:correlation_values}
\end{table*}

\floatstyle{boxed}
\restylefloat{figure}
\begin{figure*}[htp!]
\centering
%\tiny
\begin{minipage}{0.9\textwidth}
\begin{verbatim}
TRAINING_CLASSIFIER_PROMPT = """[INST]What is the topic of the following text?
\nSentence:{sentence}\nClass:{label}[/INST]"""
INFERENCE_CLASSIFIER_PROMPT = """[INST] Classify the topic of the following 
sentence.
Choose from one of the following options:{allowed_labels}.
Sentence:
{sentence}
[/INST]
Class:"""
\end{verbatim}
\end{minipage}
\caption{Instruction-tuning prompt for topic classification task (Appendix \ref{app:prompts})}\label{fig:prompt2}
\end{figure*}
\floatstyle{boxed}
\restylefloat{figure}
\begin{figure*}[htp!]
\centering
%\tiny
\begin{minipage}{0.9\textwidth}
\begin{verbatim}

TRAINING_CLASSIFIER_PROMPT = """[INST]Extract the answer of the question from the 
given context
\nQuestion:{sentence}\nContext:{context}\nAnswer:{label}[/INST]"""
INFERENCE_CLASSIFIER_PROMPT  = """[INST] Answer the question based on 
given context. Output from the given context only as in extractive QA.
Question:
{sentence}
Context:{context}
[/INST]
Answer: """
\end{verbatim}
\end{minipage}
\caption{Instruction-tuning prompt for EQA task (Appendix \ref{app:prompts})}\label{fig:prompt3}
\end{figure*}

%%%%%%%%%%%%%%%%%%%%%%%%%%%%%%%%%%%%%%%%%%%%%%%%%%%%%%%%%%%%%%%%%%%%%%%%%%%%%%%%%%%%%%
\section{Vocabulary analysis details}\label{app:analysis}
%\subsection{Performance analysis details}\label{app:perf}
\begin{figure}
    \centering
    \includegraphics[width=1.0\linewidth]{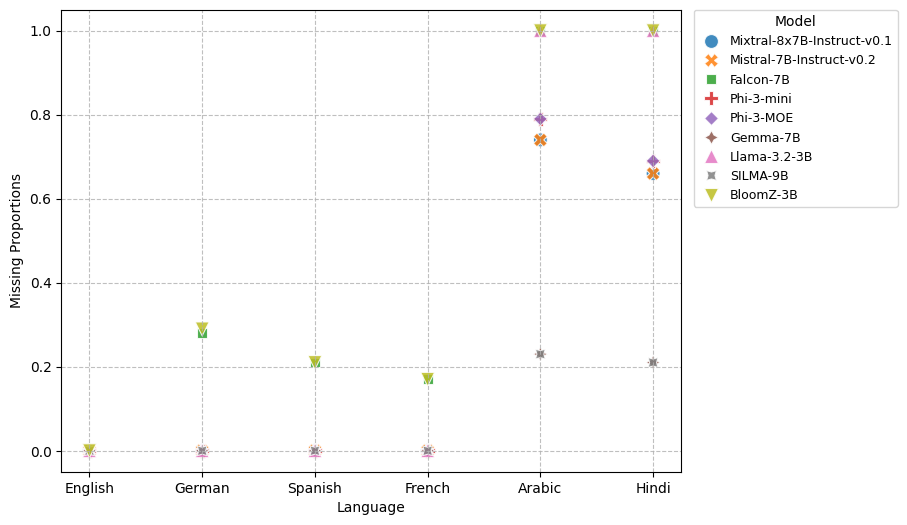}
    \caption{Missing proportions of language characters in decoder-only tokenizer vocabulary (Appendix \ref{app:vocab_uni})}
    \label{fig:LLM-miss}
\end{figure}
\begin{figure}
    \centering
    \includegraphics[width=1.0\linewidth]{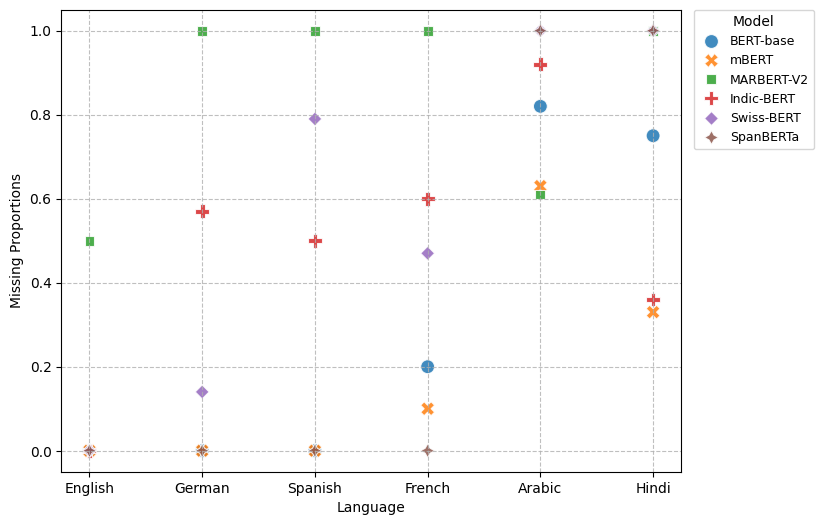}
    \caption{Missing proportions of language characters in encoder-only tokenizer vocabulary (Appendix \ref{app:vocab_uni})}
    \label{fig:encoder-miss}
\end{figure}

\begin{table*}[ht]
\centering
\renewcommand{\arraystretch}{0.9}
\begin{tabularx}{\linewidth}{l*{6}{>{\centering\arraybackslash}X}}
\hline
{} & \textbf{English} & \textbf{Arabic} & \textbf{German} & \textbf{Spanish} & \textbf{French} & \textbf{Hindi} \\
\hline
 & & & \textbf{Encoder-only} & & & \\
\hline
BERT-base & \Checkmark & \XSolidBrush & \XSolidBrush & \XSolidBrush & \XSolidBrush & \XSolidBrush \\
BERT-base-uncased & \Checkmark & \XSolidBrush & \XSolidBrush & \XSolidBrush & \XSolidBrush & \XSolidBrush \\
mBERT & \Checkmark & \Checkmark & \XSolidBrush & \Checkmark & \Checkmark & \Checkmark \\
MARBERT-V2 & \XSolidBrush & \Checkmark & \XSolidBrush & \XSolidBrush & \XSolidBrush & \XSolidBrush \\
Indic-Transformers & \Checkmark & \XSolidBrush & \XSolidBrush & \XSolidBrush & \XSolidBrush & \Checkmark \\
Swiss-BERT & \XSolidBrush & \XSolidBrush & \Checkmark & \XSolidBrush & \Checkmark & \XSolidBrush \\
SpanBERTa & \XSolidBrush & \XSolidBrush & \XSolidBrush & \Checkmark & \XSolidBrush & \XSolidBrush \\
CamemBERT & \XSolidBrush & \XSolidBrush & \XSolidBrush & \XSolidBrush & \Checkmark & \XSolidBrush \\
\hline
 & & & \textbf{Decoder-only} & & & \\
\hline
Mixtral-8x7B-Instruct-v0.1 & \Checkmark & \XSolidBrush & \XSolidBrush & \Checkmark & \Checkmark & \XSolidBrush \\
Mistral-7B-Instruct-v0.2 & \Checkmark & \XSolidBrush & \XSolidBrush & \XSolidBrush & \XSolidBrush & \XSolidBrush \\
Falcon-7B & \Checkmark & \XSolidBrush & \XSolidBrush & \XSolidBrush & \XSolidBrush & \XSolidBrush \\
phi3-mini & \Checkmark & \Checkmark & \XSolidBrush & \Checkmark & \Checkmark & \XSolidBrush \\
phi3-MOE & \Checkmark & \Checkmark & \XSolidBrush & \Checkmark & \Checkmark & \XSolidBrush \\
Gemma-7B & \Checkmark & \Checkmark & \XSolidBrush & \Checkmark & \Checkmark & \Checkmark \\
Llama3.2-3B & \Checkmark & \XSolidBrush & \XSolidBrush & \Checkmark & \Checkmark & \Checkmark \\
SILMA-9B & \Checkmark & \Checkmark & \XSolidBrush & \XSolidBrush & \XSolidBrush & \XSolidBrush \\
\hline
\end{tabularx}
\caption{\label{tab:support} Language support details of LLMs (Appendix \ref{app:lang_supp})}
\end{table*}

\subsection{Language support details}\label{app:lang_supp}

Table \ref{tab:support} presents the language support details of the decoder-only LLMs. The information is based on the support claims of each model from their respective HuggingFace pages. When a model claims \emph{"support"}, it may often refer to some representation of the language in its training data and the ability to generate or understand basic text in that language under ideal tokenization conditions. It may not guarantee robust handling of the language's words, characters, or scripts.
\begin{figure}
    \centering
    \includegraphics[width=1\linewidth]{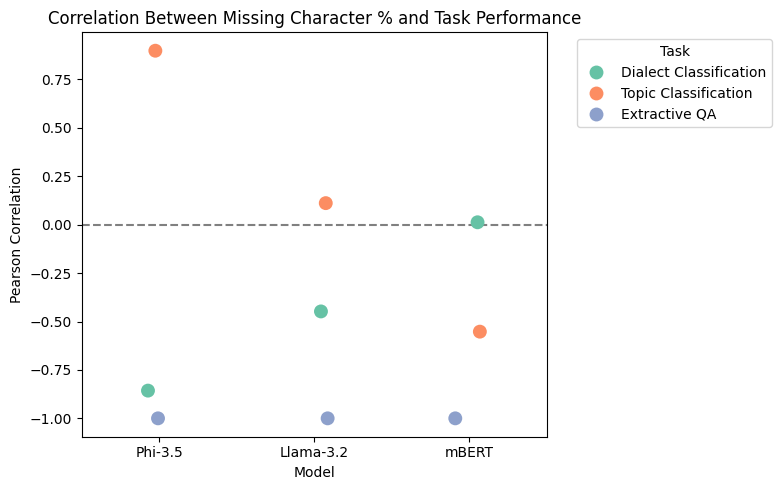}
    \caption{Correlation of missing character proportions to dialectal tasks}
    \label{fig:miss-corr-big}
\end{figure}
\subsection{Missing character proportions}\label{app:vocab_uni}

In this experiment, we compute the percentage of missing characters—those not represented as standalone tokens—in the vocabulary of each LLM. This analysis is limited to high-resource languages such as English, German, Spanish, French, Arabic, and Hindi across various LLMs. We aim to investigate potential correlations between character-level coverage and performance on dialectal downstream tasks. Although character-level analysis may initially seem counterintuitive given that most LLMs employ subword tokenizers, it sometimes becomes relevant due to their reliance on byte-level fallback mechanisms. Our qualitative analysis reveals that this fallback can sometimes negatively impact non-Latin scripts.

The Unicode ranges of the main character set used and the special characters are given in Table \ref{tab:uni}. 
% For instance, in the BloomZ and Falcon-7B models, the special Spanish characters ['ó,' 'ù,' 'û,' 'ü,' 'À'] were missing. 
As shown in Figure \ref{fig:LLM-miss}, all LLMs exhibit nearly complete character coverage for English, with a missing proportion of 0.0 (lower is better). In contrast, the missing proportion for non-Latin languages is considerably higher than for Latin languages across most multilingual decoder-based models (e.g., LLaMA, Phi, Mixtral), with the exception of the Gemma model. Among language-specific decoder models, the missing proportion is notably lower—for instance, SILMA reports only 23\% missing characters in Arabic.

For encoder-only models (see Figure \ref{fig:encoder-miss}), language-specific encoders tend to achieve better character coverage in their respective languages. Notably, mBERT maintains reasonable coverage over non-Latin scripts. However, MARBERT exhibits a substantial proportion of missing Arabic characters at the token level. This is likely due to its frequency-based subword tokenizer, where individual characters are often absorbed into larger subword units.

While such missing character coverage does not necessarily impair performance in language-specific models—owing to their strong modeling of linguistic structure across granularities—it can pose challenges for broader multilingual LLMs. Ensuring at least character-level granularity in these models may help mitigate issues arising from multibyte representations of non-Latin scripts.

% \textbf{Task Analysis:} It was noted that the SILMA model, which is based on Gemma2, gives a reasonable \textbf{parity} of 1.24 with Hindi and a \textbf{missing proportion} of only 21\%. While with Phi3, the missing proportion is about 69\%, with a parity of 4.59. With the Llama model, the parity is 2.53, while the missing proportion was 100\%\footnote{Hindi character-level tokens were absent}. This aligns with its performance in the ILI dataset, where the Devanagari script seems underrepresented. Regarding parity, Llama3 seems to be better than Phi3, which could be related to the limitation of the parity metric discussed in Section \ref{sec:qual}. It was observed that Phi3 models usually interpret Hindi text at a character level with its fall-back approach and 31\% representation of the characters in vocabulary. From Table \ref{tab:Results}, it can also be observed that Llama performance in all non-Latin script datasets (NADI, ILI) is relatively less compared to the phi3-mini (Note Llama model doesn't claim Arabic support).\\
% In Figure \ref{fig:miss}, we depict the proportion of missing characters for each LLM across the languages. For each language, we use their basic character sets; for Latin scripts, we also include special characters, if any (details in Appendix \ref{app:vocab_uni}).   
We used the three models for correlation analysis - mBERT, Phi-3.5, and Llama-3.2. From Figure \ref{fig:miss-corr-big}, it can be observed that negative correlations dominate, especially for Phi-3.5 and Llama-3.2. All models show high negative correlations with EQA task, indicating that higher character coverage (fewer missing characters) improves performance. In TC, both decoder-only LLMs show positive correlations.